\pdfoutput=1
\documentclass[12pt]{article}
\usepackage[letterpaper,margin=.77in]{geometry}

\usepackage[utf8]{inputenc}
\usepackage[T1]{fontenc}
\usepackage{amsmath}
\usepackage{amssymb}
\usepackage{amsthm}
\usepackage{graphicx}
\usepackage{xcolor}
\usepackage{numprint}
\PassOptionsToPackage{hyphens}{url}
\usepackage{hyperref}

\makeatletter
\renewcommand\paragraph{\@startsection{paragraph}{4}{\parindent}%
  {0.8ex \@plus.3ex \@minus.2ex}%
  {-1em}%
  {\normalfont\normalsize\bfseries}}
\renewcommand\section{\@startsection{section}{1}{\z@}%
  {-3.3ex \@plus -1ex \@minus -.2ex}%
  {2.1ex \@plus.2ex}%
  {\normalfont\Large\bfseries}}
\renewcommand\subsection{\@startsection{subsection}{2}{\z@}%
  {-3.1ex \@plus -1ex \@minus -.2ex}%
  {1.4ex \@plus.2ex}%
  {\normalfont\large\bfseries}}
\makeatother

\newcommand{\Oh}[1]{\mathcal{O}\!\left( #1\right)}

\newcommand{\etal}{et~al.~}

\usepackage{tikz,pgfplots}
\usepgfplotslibrary{colorbrewer}
\usetikzlibrary{patterns}
\npdecimalsign{.}

\usepackage{float}
\usepackage{algpseudocode}
\usepackage{algorithm}
\usepackage{booktabs}
\usepackage{longtable}

\algnewcommand\algorithmicinput{\textbf{Input:}}
\algnewcommand\INPUT{\item[\algorithmicinput]}
\algnewcommand\algorithmicoutput{\textbf{Output:}}
\algnewcommand\OUTPUT{\item[\algorithmicoutput]}

\newcommand{\CC}{C\texttt{++}}

\newcommand{\gmseqbase}{25.0}
\newcommand{\gmseqtuned}{7.28}

\newcommand{\spseq}{3.44}

\newcommand{\peakseqbase}{492}
\newcommand{\peakseqtuned}{193}

\newcommand{\ncommon}{49}
\newcommand{\nparbasefail}{4}
\newcommand{\gmcomparbase}{23.0}
\newcommand{\gmcompartuned}{1.07}
\newcommand{\spparcom}{21.49}

\newcommand{\nkcore}{20}
\newcommand{\spseqkcore}{1.28}
\newcommand{\spparkcore}{1.63}
\newcommand{\spseqgen}{6.26}
\newcommand{\sppargen}{127}
\newcommand{\gmcomparmid}{3.65}
\newcommand{\spparsessionone}{6.29}
\newcommand{\spparsessiontwo}{3.42}

\pgfplotscreateplotcyclelist{black white 2}{%
	every mark/.append style={solid},mark=*\\%
	every mark/.append style={solid},mark=square*\\%
	every mark/.append style={solid},mark=o\\%
	mark=star\\%
	every mark/.append style={solid},mark=diamond*\\%
	densely dashed,every mark/.append style={solid},mark=*\\%
	densely dashed,every mark/.append style={solid},mark=square*\\%
	densely dashed,every mark/.append style={solid},mark=o\\%
	densely dashed,every mark/.append style={solid},mark=star\\%
	densely dashed,every mark/.append style={solid},mark=diamond*\\%
}

\pgfplotsset{
  cycle list/Dark2,
  every axis/.append style={
    ylabel near ticks,
    log basis y={2},
    log basis x={2},
    legend cell align={left},
    legend style={font=\Large},
    label style={font=\Large},
    title style={font=\Large},
    tick label style={font=\Large},
    cycle multiindex* list={
      Dark2
      \nextlist
      black white 2
      \nextlist
   },
  }
}

\def\MdN{\ensuremath{\mathbb{N}}}

\title{Agentic Algorithm Engineering: Improving Shared-Memory Exact Minimum Cuts}

\newcounter{heidelbergmark}

\author{
David A. Bader\thanks{New Jersey Institute of Technology, Newark, NJ, USA}
\and
Adil Chhabra\thanks{Heidelberg University, Heidelberg, Germany. Christian Schulz is the corresponding author.}%
\setcounter{heidelbergmark}{\value{footnote}}
\and
Ernestine Gro{\ss}mann\footnotemark[\value{heidelbergmark}]
\and
Monika Henzinger\thanks{Institute of Science and Technology Austria (ISTA), Klosterneuburg, Austria}
\and
Alexander Noe\thanks{Amazon.com, New York, NY, USA. Work done prior to joining Amazon.}
\and
Christian Schulz\footnotemark[\value{heidelbergmark}]
}

\date{}

\begin{document}

\maketitle

        \vspace*{-1cm}
\begin{abstract}
The minimum cut problem for an undirected edge-weighted graph asks us to divide its set of nodes into two blocks while minimizing the weighted sum of the cut edges. Over the last years, we engineered a range of fast algorithms for this problem. Our fastest \emph{exact} algorithm uses an \emph{inexact} algorithm to obtain a better bound for the problem, reductions that depend on this bound, improved data structures and parallel contraction routines. It is available in the open-source package \texttt{VieCut} and, on real-world instances, outperformed the previously fastest solvers by a factor of up to $2.5$ sequentially and up to $12.9$ when run in parallel. 

We improve this algorithm using \emph{agentic algorithm engineering (AAE)}, a methodology that we introduce here, in which autonomous large language model agents run the algorithm engineering cycle on an existing code base: they form hypotheses about where running time is lost, implement them, benchmark the result on a fixed instance set and keep or discard the change. Even though we had already tuned our algorithm by hand extensively, the agent finds significant optimizations, in particular on the DIMACS core instances: factors of $\spseqkcore$ (sequential) and $\spparkcore$ ($32$ threads) on real-world $k$-cores, and $\spseqgen$ and $\sppargen$ on the \hbox{DIMACS core instances.}
\end{abstract}

\vspace*{-.6cm}
\section{Introduction}

Given an undirected graph with non-negative edge weights, the \emph{minimum cut problem} asks for a partition of the vertices into two sets that minimizes the total weight of the edges between them; its value is also called the \textit{edge connectivity} of the graph~\cite{nagamochi1992computing,henzinger2017local}. Minimum cuts are used in network reliability~\cite{karger2001randomized,ramanathan1987counting}, VLSI design~\cite{krishnamurthy1984improved} and as a subroutine in branch-and-cut algorithms for the traveling salesman and other combinatorial problems~\cite{padberg1991branch}, so they have to be computed quickly on large inputs. All exact algorithms have non-linear running time, and those of Hao~\etal\cite{hao1992faster} and Nagamochi~\etal\cite{nagamochi1992computing,nagamochi1994implementing}, which are orders of magnitude faster than the others in practice~\cite{Chekuri:1997:ESM:314161.314315,henzinger2018practical,junger2000practical}, do not parallelize easily. Appendix~\ref{app:related} surveys related work.

\looseness=-1 Over the last years, we engineered a range of algorithms for the minimum cut problem and its variants. We first gave a practical shared-memory parallel \emph{heuristic} algorithm, available in the open-source package \texttt{VieCut}~\cite{henzinger2018practical,henzinger2018practicaljv}, which computes near-optimal and often exact minimum cuts very quickly. We then used this algorithm to obtain a better bound $\hat\lambda$ and engineered a shared-memory parallel \emph{exact} algorithm~\cite{henzinger2018shared} on top of it. It outperformed the previously fastest solvers, the implementations of the algorithms of Nagamochi~\etal and of Hao and Orlin, by a factor of up to $2.5$ sequentially and up to $12.9$ when run in parallel on real-world instances. 

Recently, large language models (LLMs) have been used not only to generate code, but also to discover and improve algorithms~\cite{romeraparedes2024funsearch,novikov2025alphaevolve}. This includes low-level optimizations that so far required a large amount of manual work. We use the algorithm engineering methodology as a starting point and introduce \emph{agentic algorithm engineering}, a methodology in which autonomous LLM agents run the algorithm engineering cycle on an existing code base. The agent receives the code and a fixed set of benchmark instances, and modifies the code to make it faster on the whole set rather than on a single instance. {Unlike classical auto-tuning~\cite{ansel2014opentuner}, which searches a fixed parameter space of an unchanged code, the agent modifies the code itself, reaching changes without a tuning knob, such as replacing a data structure or repartitioning \hbox{parallel work}.}

\textbf{Contribution.}
\emph{First}, we summarize our shared-memory parallel \emph{exact} minimum cut algorithm~\cite{henzinger2018shared} in a self-contained way.
\emph{Second}, we introduce \emph{agentic algorithm engineering} (AAE), in which an autonomous LLM agent runs the algorithm engineering cycle on an existing code base: it inspects the code and the experimental infrastructure, forms hypotheses about where running time is lost, implements candidate optimizations and benchmarks them on a fixed set of instances. Improvements are kept as the starting point of the next round; the rest is discarded together with the reason for its failure, so that later iterations can build on it.
\emph{Third}, we run the agent, Claude Opus~5~\cite{anthropic2026opus5}, on our exact algorithm on the challenge instances and on the instances of the original publication. It finds further optimizations, factors of $\spseqkcore$ sequentially and $\spparkcore$ at $32$ threads on the real-world $k$-cores, and $\spseqgen$ and $\sppargen$ on the DIMACS core instances, for about \$$350$ in API usage ($1.1$ million output tokens). The agent tunes on the same $53$ instances on which we report these speedups; we ask how much an agent gains by tuning a hand-optimized code to a given workload, not how the result fares on unseen instances (Section~\ref{ss:sessions}).

\vspace*{-0.25cm}
\section{Preliminaries}\label{s:preliminaries}
\label{prelim}

\paragraph{Basic Concepts.}
\looseness=-1 Let $G = (V, E, c)$ be an undirected graph with $n = |V|$ vertices, $m = |E|$ edges and non-negative edge weights $c: E \rightarrow \MdN$, extended to edge sets by summation. The \emph{degree} of a vertex is the total weight of its incident edges. A \emph{cut} $(A, V \setminus A)$ partitions $V$ into two non-empty \emph{sides}; its \emph{capacity} is the total weight of the edges between them. A \emph{minimum cut} has smallest capacity, denoted by $\lambda(G)$ or simply $\lambda$, and the \emph{minimum $s$-$t$-cut} $\lambda(G,s,t)$ is the smallest cut that separates the vertices $s$ and $t$. $\hat\lambda$ denotes the smallest upper bound on $\lambda$ discovered so far; since the \emph{trivial cut} $(\{u\}, V\setminus \{u\})$ has capacity equal to the degree of $u$, the minimum degree serves as an initial bound. \emph{Contracting} an edge $\{u, v\}$ merges $u$ and $v$ into one vertex that inherits their other edges, summing the weights of edges that become parallel.

\paragraph{Related Work.}
\label{related}

Exact minimum cut algorithms are either flow-based, computing the global minimum cut from maximum flow computations~\cite{ford1956maximal,gomory1961multi,hao1992faster}, or contraction-based, shrinking the graph while preserving at least one minimum cut~\cite{padberg1990efficient,nagamochi1992computing,nagamochi1994implementing,karger1996new}. Recent theoretical work has lowered both the randomized and the deterministic bounds to near-linear~\cite{gawrychowski2020minimum,ghaffari2020faster,li2021deterministic,henzinger2024deterministic,anderson2023parallel}, but none of these algorithms have been implemented. Appendix~\ref{app:related} gives a detailed overview.
Our \texttt{VieCut} algorithm is available as the open-source package \texttt{VieCut}~{\cite{viecut-code}} as well as in the \texttt{CHSZLabLib}~{\cite{chszlablib}} via an easy-to-use Python interface. Since its publication it has been used as a minimum cut engine in a range of applications such as well-connected community detection~\cite{park2024wellconnected,ramavarapu2024cmpp,dindoost2025wellconnected}, weighted connectivity augmentation~\cite{faraj2024augmentation}, analysis of coalitional games~\cite{barr2024coalitional} or in data reduction rules~\cite{schulz2022cluster}.

\subsection{The Exact Minimum Cut Algorithm \texttt{VieCut}}
\label{cuts}

\looseness=-1 To be self-contained, we summarize the main components of our shared-memory parallel \emph{exact} minimum cut algorithm~\cite{henzinger2018shared}, which is the algorithm that we later optimize using agentic algorithm engineering. Everything described in this section is prior work.
The algorithm is based on the contraction-based exact algorithm of Nagamochi~\etal\cite{nagamochi1992computing,nagamochi1994implementing}, which we call \texttt{NOI}. Their algorithm uses a routine called CAPFOREST, which traverses the graph in a modified breadth-first order and computes a lower bound $q(e)$ of the connectivity $\lambda(G,u,v)$ for each edge $e=\{u,v\}$. If the connectivity between two vertices is at least as large as the current upper bound for the minimum cut, then the edge between them can be contracted. Hence, edges with $q(e) \geq \hat\lambda$ can be safely contracted, and the algorithm is guaranteed to find at least one such edge per run. This is repeated until only two vertices remain. The running time therefore depends on the bound $\hat\lambda$, which determines how many edges can be contracted, and on the cost of the priority queue operations in CAPFOREST; \texttt{VieCut} improved both and parallelized the routine.

\paragraph*{Lowering the upper bound.} The algorithm first runs the \emph{heuristic} shared-memory parallel algorithm to obtain a better upper bound $\hat\lambda$ than the minimum degree. This heuristic~\cite{henzinger2018practical} repeatedly computes a clustering with parallel label propagation~\cite{raghavan2007near}, contracts each cluster into a single vertex and applies the local contraction tests of Padberg and Rinaldi~\cite{padberg1990efficient}; once only a constant number of vertices remain, the contracted graph is solved exactly with \texttt{NOI}. The resulting cut is a cut of the input graph and thus a valid upper bound, but a cluster may contain vertices from both sides of a minimum cut, so it need not be optimal.
In practice the heuristic algorithm almost always returns the exact minimum cut. As the contraction tests depend on $\hat\lambda$, a smaller bound allows more edges to be contracted in each round. This shrinks the graph faster and lowers the number of CAPFOREST runs.

\paragraph*{Bounded priority queues.} CAPFOREST uses a priority queue $\mathcal{Q}$ in which the key of a vertex is its connectivity to the already scanned vertices. We showed that the algorithm remains correct if the priorities in the queue are limited to $\hat\lambda$, meaning that elements in the queue having a key larger than $\hat\lambda$ are not updated. {As the priorities are then integers bounded by $\hat\lambda$, $\mathcal{Q}$ can be implemented as a bucket priority queue with one bucket per attainable key; the implementation offers two bucket variants (FIFO and LIFO bucket order) alongside a binary heap.}

\paragraph*{Parallel CAPFOREST and contraction.} We then adapted the algorithm so that contractible edges can be detected in parallel (Algorithm~\ref{algo:parnoi}). Every process starts at a random vertex and runs a modified CAPFOREST, where every vertex is visited by only one process. This way, no process has to scan the whole graph, while the vertices in sparse regions of the graph, which might otherwise not be scanned by any process, are still scanned. We showed that the values $q(e)$ are still valid lower bounds, so that no cut smaller than $\hat\lambda$ is lost. Contraction is \hbox{parallelized as well.}

\paragraph*{Overall algorithm.} Lastly, everything is put together (Algorithm~\ref{algo:parmc}). The algorithm first runs the \emph{heuristic} available in \texttt{VieCut} to obtain $\hat\lambda$ and then alternates parallel CAPFOREST and parallel graph contraction until only two vertices are left. In the unlikely case that no contractible edge was found, CAPFOREST is run sequentially. Contraction merges several vertices of the input graph into a single vertex of the contracted graph, which we call a \emph{collapsed} vertex. The degree of a collapsed vertex is the capacity of the cut that separates the vertices it represents from the rest of the graph, and is therefore a valid upper bound on the minimum cut. Whenever we encounter a collapsed vertex whose degree is smaller than $\hat\lambda$, we can thus update the upper bound.

\vspace*{-0.25cm}
\section{Agentic Algorithm Engineering}
\label{s:agentic}

\looseness=-1 \emph{Algorithm engineering}, as described by Sanders~\cite{sanders2009algorithm}, organizes design, analysis, implementation and experimental evaluation into a cycle instead of a one-directional pipeline. Realistic machine models and real-world inputs are used instead of idealized assumptions, and experiments test falsifiable hypotheses about an implementation. Their outcome then drives the next round of design decisions. Reusable implementations are outcomes of this process as well.

\paragraph*{Overview.} \looseness=-1 In \emph{agentic algorithm engineering} (AAE), an autonomous LLM agent runs this cycle.\footnote{\url{https://github.com/CHSZLab/AgenticAlgorithmEngineering}} A session is given four inputs: a \emph{target program} whose source the agent may modify, a \emph{metric} to be improved,\footnote{In this paper, since we have an exact algorithm, we optimize for running time; however the metric could also be solution quality of a heuristic algorithm or memory footprint of an algorithm.} a \emph{benchmark} that produces reproducible measurements and a \emph{time budget} per experiment. The agent then runs an endless loop in which each iteration is one pass through the cycle. The first experiment of a session is always the baseline, so that all later results are relative to a measurement taken with the same benchmark on the same machine. All changes are kept under version control, so that experiments can be traced and reverted. Since an agent that optimizes for running time alone can easily produce code that is fast but wrong, every change is checked against a set of correctness assertions before it is benchmarked. The agent derives these assertions from the code, and the user can supply further ones.

\paragraph*{The cycle.} Each iteration has six steps. In the \emph{model} step, the agent maintains an explicit picture of the hot paths, the hardware and where time is spent, and updates it whenever a measurement contradicts it. In the \emph{design} step, it formulates a falsifiable hypothesis of the form ``changing $X$ improves $Y$ by roughly $Z$, because \ldots'', grounded in the current picture or in earlier results and changing one thing at a time. In the \emph{analysis} step, it predicts direction and magnitude of the effect, which filters out experiments not worth running. In the \emph{implementation} step, it applies a minimal change and commits it, so that every experiment is exactly one git commit. In the \emph{experiment} step, it runs the benchmark, with independent runs concurrently if specified, and aggregates the running times over the instances by the geometric mean. In the \emph{evaluation} step, it compares the result to the current best: improvements are kept and become the new reference point, regressions are reverted, crashes are repaired or given up on, and a small improvement that adds much complexity is not kept.
The agent may change everything inside the target files: data structures, parameters, memory layout, control flow, parallelization, up to replacing a subcomponent by a different algorithm. It may not change the benchmark, the metric, the measurement methodology, the input data or the time budget, and it may not add dependencies. These constraints ensure that the agent can only improve the specified metric.

\paragraph*{Bookkeeping.} Every experiment, including the failed ones, is logged with its commit hash, the metric value, the resource usage, the decision and the hypothesis that motivated it. A session is therefore fully documented. In addition, a community-maintained knowledge base of earlier sessions can be consulted when generating hypotheses. The sessions in this paper did not use it; they started from the \hbox{code base alone.} In particular, the agent is not seeded with a curated list of optimization ideas: the hypotheses come from the model's own knowledge of algorithm and performance engineering, combined with what it observes in the code, the profiles and the measurements.

\section{Experiments and Results}
\label{impl}

\paragraph*{Baseline.} Our baseline is the exact \texttt{VieCut} algorithm~\cite{henzinger2018shared}, used unchanged; its original evaluation, on a different machine and on real-world instances only, is summarized in the introduction.

\paragraph*{Setup/Instances.} All experiments, including the benchmark runs of the agent sessions, run on a machine with an AMD EPYC 7702P processor and $1$\,TiB of main memory in a single NUMA domain, running Ubuntu 24.04. All code is written in \CC{}, compiled with \texttt{g++}~13.3 using \texttt{-O3 -march=native}, and uses OpenMP for shared-memory parallelism.
The benchmark has $53$ instances. The $33$ generated ones, three from each of eleven families, come from the generators of the DIMACS challenge, which are described by Chekuri~\etal\cite{Chekuri:1997:ESM:314161.314315} and Levine~\cite{levine1997experimental} but were never published, so we reimplemented them from these descriptions; as the challenge designates the families of Chekuri~\etal as its core benchmark, we call them the \emph{DIMACS core instances}. The $20$ real-world ones are four $k$-cores of each of the five web and social graphs of the original evaluation~\cite{henzinger2018shared}. Appendix~\ref{app:instances} lists every instance.

\subsection{Agentic Optimization of the Exact Algorithm}
\label{ss:sessions}

We instantiate the methodology of Section~\ref{s:agentic} for \texttt{VieCut} in two sessions: a single-threaded session starting from the unmodified implementation, and a session at $32$ threads starting from the result of the first. Together they comprise $44$ experiments, of which $23$ were accepted. An experiment is accepted only if the geometric mean of the running time over the $53$ instances improves, every instance runs to completion, and every instance reports the same minimum cut as the baseline. Each experiment times every instance once; the acceptance statistic is the geometric mean over all $53$ instances, so a small factor such as the $1.027$ below, which would be within run-to-run variability on a single instance, is accepted only because it improves the whole set.

\paragraph*{Agent setup.} \looseness=-1 The agent is Claude with the Opus~5 model~\cite{anthropic2026opus5} (Anthropic), driven by the execution loop of Section~\ref{s:agentic}. Session state lives on disk, not in the model's context window, which is finite and volatile: one benchmark run, build or profile prints far more than the agent needs, and a long conversation is compacted. Output is therefore redirected to log files, and the agent reads back only a few extracted lines per run, such as the geometric mean of the running time and the result of the correctness checks. What to read back is the design decision, since the model can only reason about what is in its context: too much buries the signal and costs tokens on every iteration, too little loses the experience of earlier experiments. The durable record is a ledger with one row per experiment, holding its hypothesis and outcome, one git commit per experiment and file snapshots for reverting refuted changes; a compaction of the conversation therefore loses nothing essential. The bottleneck of a session is the evaluation time of the $53$-instance benchmark, not the agent's reasoning or \hbox{implementation time.}

\begin{figure}[t]
\centering
\begin{tikzpicture}
\begin{loglogaxis}[
    width=0.47\textwidth, height=5.0cm,
    title={one thread}, title style={font=\small, yshift=-3pt},
    xlabel={VieCut running time [s]},
    ylabel={AAE-optimized running time [s]},
    xmin=0.1, xmax=10000, ymin=0.1, ymax=10000,
    log basis x={10}, log basis y={10},
    legend pos=north west, legend cell align={left},
    legend style={font=\footnotesize, draw=gray!50, fill=white, fill opacity=0.9, text opacity=1},
    label style={font=\small}, tick label style={font=\small},
    grid=both, major grid style={gray!25, very thin},
    minor grid style={gray!12, very thin},
    axis line style={gray!60},
]
\addplot[gray!70, thin, domain=0.1:10000, samples=2, forget plot] {x};
\addplot[gray!55, densely dashed, domain=1:10000, samples=2, forget plot] {x/10};
\addplot[gray!45, densely dotted, domain=10:10000, samples=2, forget plot] {x/100};
\addplot[mark=*, mark size=1.5pt, only marks, draw=teal!80!black, fill=teal!45, thick] coordinates {(1.82722,0.418572) (3.24017,0.763141) (4.69376,0.970392) (5.68572,1.41811) (6.73369,1.90469) (8.1622,3.83019) (9.24399,8.26472) (9.97005,2.43375) (11.2521,4.58099) (11.6518,4.76078) (12.0555,3.72807) (14.5724,0.385369) (15.6547,4.14157) (21.652,5.44957) (34.1186,2.00557) (36.42,28.6842) (41.7253,4.26314) (43.1026,25.4803) (43.1477,2.66216) (75.7042,0.532129) (87.5828,22.7187) (98.5264,23.9108) (123.154,0.913097) (164.13,14.8339) (198.345,48.5755) (209.413,57.0141) (210.257,59.4148) (211.715,52.1174) (215.121,1.74809) (228.587,47.1958) (468.046,132.545) (939.292,68.5295) (3225.62,283.446)};
\addlegendentry{DIMACS core instances}
\addplot[mark=triangle*, mark size=2.1pt, only marks, draw=orange!85!black, fill=orange!55, thick] coordinates {(0.462283,0.34662) (1.27418,0.979373) (1.35238,0.979659) (1.48165,1.06938) (2.33017,1.75158) (2.3368,1.82722) (4.7443,3.66118) (5.13961,4.05013) (5.73528,4.41691) (6.10413,4.2957) (6.25437,4.74477) (13.7346,10.0249) (18.7694,15.8846) (30.0534,22.9423) (62.2002,54.0393) (68.274,52.521) (74.3058,64.37) (96.7104,87.6858) (108.588,97.5599) (145.651,108.605)};
\addlegendentry{real-world $k$-cores}
\end{loglogaxis}
\end{tikzpicture}
\hfill
\begin{tikzpicture}
\begin{loglogaxis}[
    width=0.47\textwidth, height=5.0cm,
    title={$32$ threads}, title style={font=\small, yshift=-3pt},
    xlabel={VieCut running time [s]},
    ylabel={},
    xmin=0.01, xmax=10000, ymin=0.01, ymax=10000,
    log basis x={10}, log basis y={10},
    legend pos=north west, legend cell align={left},
    legend style={font=\footnotesize, draw=gray!50, fill=white, fill opacity=0.9, text opacity=1},
    label style={font=\small}, tick label style={font=\small},
    grid=both, major grid style={gray!25, very thin},
    minor grid style={gray!12, very thin},
    axis line style={gray!60},
]
\addplot[gray!70, thin, domain=0.01:10000, samples=2, forget plot] {x};
\addplot[gray!55, densely dashed, domain=0.1:10000, samples=2, forget plot] {x/10};
\addplot[gray!45, densely dotted, domain=1:10000, samples=2, forget plot] {x/100};
\addplot[mark=*, mark size=1.5pt, only marks, draw=teal!80!black, fill=teal!45, thick] coordinates {(2.07774,0.0370851) (3.54932,0.056448) (6.174,0.0693128) (13.863,2.59313) (16.6789,1.22687) (23.0005,2.621) (23.125,0.180305) (49.3623,0.214598) (85.1749,1.70521) (93.2801,0.592797) (95.2337,1.87548) (142.805,0.999255) (163.08,5.28864) (181.642,2.27347) (184.69,0.297754) (198.604,3.04404) (198.751,4.49345) (199.68,3.01678) (200.439,3.01101) (232.205,2.79442) (251.067,2.09254) (257.099,0.0597491) (406.507,0.0755031) (431.563,5.25292) (527.502,1.10877) (1290.42,1.09577) (1713.11,2.15362) (1801.93,38.7054) (3568.63,0.897184)};
\addlegendentry{DIMACS core instances}
\addplot[mark=triangle*, mark size=2.1pt, only marks, draw=orange!85!black, fill=orange!55, thick] coordinates {(0.0590172,0.214513) (0.224223,0.198969) (0.242674,0.18315) (0.493702,0.329011) (0.580176,0.404815) (0.590089,0.346207) (0.600187,0.426004) (1.69753,1.2416) (2.47209,0.998772) (2.98695,1.91754) (3.25604,0.526152) (3.49686,3.3535) (4.02071,0.562509) (6.44423,0.612472) (6.96402,4.31782) (7.8303,6.30409) (8.97823,7.47932) (9.59763,9.38126) (12.0953,11.9278) (14.3278,10.5087)};
\addlegendentry{real-world $k$-cores}
\end{loglogaxis}
\end{tikzpicture}
\caption{Running time of the AAE-optimized implementation against the unmodified one, one point per instance; the guides mark equality and speedups of $10$ and $100$. The $4$ instances the unmodified implementation cannot solve at $32$ threads are omitted from the right panel.}
\label{fig:scatter}
\end{figure}
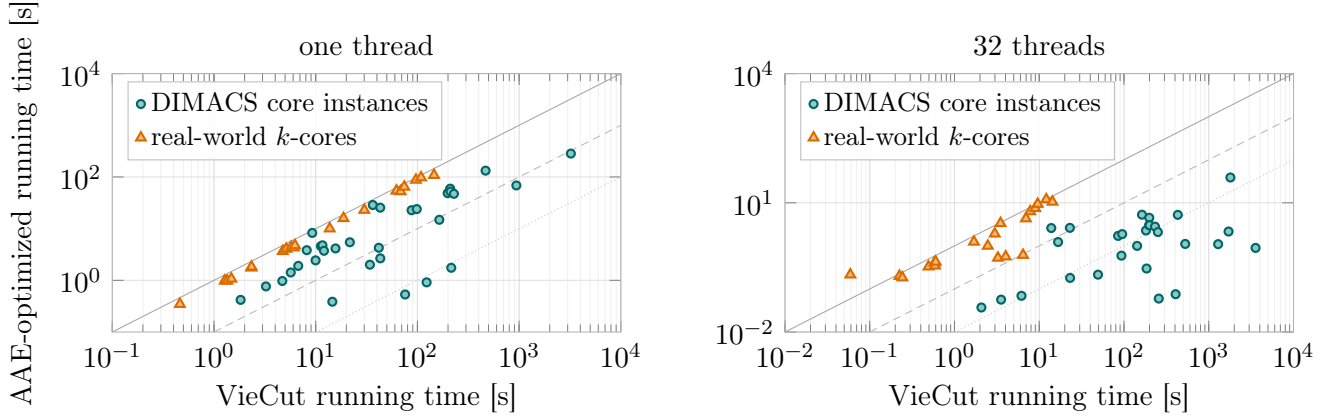

\paragraph*{The implementation in outline.} \looseness=-1 Some details of the unmodified implementation are needed below. We call one iteration of CAPFOREST and contraction in the exact phase (Section~\ref{cuts}) a \emph{round}. A round costs $\Theta(n+m)$ time, and the number of rounds is the decisive quantity for the running time: on most graphs, a round contracts a constant fraction of the vertices and $\Oh{\log n}$ rounds suffice, while on the DIMACS core instances of Appendix~\ref{app:instances}, which are constructed to resist contraction, a round merges a single vertex pair and $\Theta(n)$ rounds are needed. In the parallel algorithm, each thread runs a CAPFOREST \emph{pass} from its own random start vertex. In the unmodified implementation, the passes share the array that marks visited vertices, so a vertex taken by one pass is skipped by the others and the passes divide the traversal between them. The graph is stored as one adjacency vector per vertex with an edge record of $40$ bytes across five fields, of which CAPFOREST reads two, the target and the weight. Its inner loop runs once per directed edge per round and is the hottest path of the algorithm.
\paragraph*{Overall improvement.} \looseness=-1 On one thread, the accepted changes reduce the geometric mean of the running time over all $53$ instances from $\gmseqbase$\,s to $\gmseqtuned$\,s, a factor of $\spseq$, and the peak memory from $\peakseqbase$\,GiB to $\peakseqtuned$\,GiB. At $32$ threads, the unmodified implementation fails on $\nparbasefail$ of the $53$ instances even when granted the entire memory of the machine and two hours per instance, whereas the modified implementation solves all $53$; over the $\ncommon$ instances that all configurations solve, the accepted changes reduce the geometric mean of the running time from $\gmcomparbase$\,s to $\gmcompartuned$\,s, a factor of $\spparcom$. Figure~\ref{fig:scatter} shows the per-instance running times.

\looseness=-1 These factors have to be read against the instance set. Most DIMACS core instances are worst cases for contraction-based algorithms, where a round merges a single vertex pair so that any per-round cost is paid $\Theta(n)$ instead of $\Oh{\log n}$ times, and several have minimum cuts orders of magnitude larger than those of the original evaluation, for which data structures sized by the cut value were appropriate. Split accordingly, the accepted changes are worth factors of $\spseqkcore$ on one thread and $\spparkcore$ at $32$ threads on the $\nkcore$ real-world $k$-cores, the kind of instance the implementation was tuned on, and $\spseqgen$ and $\sppargen$ on the DIMACS core instances; all $\nparbasefail$ instances that the unmodified implementation cannot solve are DIMACS core instances.

\paragraph*{Single-threaded session.} Of the $16$ experiments of this session, $11$ were accepted. The two largest improvements remove costs that grow with {the minimum cut value and the graph size, respectively}. Every speedup quoted for an individual change below is the improvement of the geometric mean of the running time over all $53$ instances, measured by the agent against the code with all previously accepted changes. (1) CAPFOREST keys its priority queue by weights in $[0,\hat\lambda]$, so the bucket variants of the queue, which hold one bucket per attainable key, are sized by the value of the minimum cut rather than by the graph. Each bucket is a \texttt{std::deque} that allocates its first block on construction, about a kilobyte per attainable key: on a weighted instance with $8000$ vertices and a minimum cut of $1.6 \cdot 10^8$, the queue alone requires roughly $200$\,GiB, rebuilt in every round. Selecting a binary heap, whose size is $\Oh{n}$, whenever the cut value exceeds both the number of vertices and $\numprint{10000}$ yields a speedup of $\numprint{1.544}$; the instance above goes from $215$\,s and $202$\,GiB to $3.5$\,s and $2.2$\,GiB. (2) Both the sequential and the parallel exact loop rebuilt the contracted graph in every round and kept all intermediate graphs alive, a history that is only ever read when the sides of the cut are requested rather than its value. On the families where a round merges a single pair, this costs $\Theta(n+m)$ time, and one live graph, per constant amount of progress. Contracting in place, keeping a single copy of the graph, yields speedups of $\numprint{1.277}$ and $\numprint{1.216}$ in the two loops. (3) The heuristic only coarsens graphs above $\numprint{10000}$ vertices and finishes with a sequential exact solve, so {on inputs with $n < \numprint{10000}$, the value it returns is not a bound but the exact minimum cut; returning it directly yields a speedup of} $\numprint{1.093}$. The remaining accepted changes contribute one to five percent each by removing work from the inner loop: (4) flags stored as bytes instead of bit-packed \texttt{std::vector<bool>}, whose probes cost a shift and a mask; (5) one lookup of a vertex's queue position per update instead of two; (6) resolving a vertex's adjacency vector once instead of once per edge; and (7) omitting, in the contraction, an allocation per vertex per round for group lists \hbox{that are never read.}

\paragraph*{Parallel session.} \looseness=-1 This session starts from the code produced by the single-threaded session, whose changes are effective in parallel as well: at $32$ threads, that code already lowers the geometric mean of the running time over the $\ncommon$ common instances from $\gmcomparbase$\,s to $\gmcomparmid$\,s, a factor of $\spparsessionone$. The changes accepted here contribute a further factor of $\spparsessiontwo$, composing to the $\spparcom$ reported above. {Of the $28$ experiments in this session, $12$ were accepted. The five changes described below are the largest ones. (1) One source of improvement is the treatment of the visited-vertex array in concurrent CAPFOREST passes, where the better strategy turns out to be instance dependent. Sharing the array lets a pass skip vertices already visited by another thread, reducing redundant work, but fragments what each pass sees and can therefore reduce the number of contractible edges found in a round. With private arrays, each pass performs a complete traversal, increasing work but potentially finding more contractions. Neither strategy consistently dominates: private arrays are about ten times faster on the random graphs, whereas sharing is several times faster on the real-world instances. Alternating the two settings over the first rounds and retaining the better one yields a speedup of $\numprint{1.175}$.}

{(2) A second bottleneck occurs on small graphs. For instances with fewer than $\numprint{10000}$ vertices, the heuristic algorithm used by the original implementation computes the upper bound $\hat\lambda$ using a sequential solver rather than running label propagation contraction. Fifteen of the $53$ instances fall below this threshold, so this sequential step becomes a bottleneck while the remaining threads are idle. For these instances, we instead skip the heuristic upper bound computation, initialize $\hat\lambda$ with the minimum vertex degree, and proceed directly with the parallel reductions and CAPFOREST. This yields a speedup of $\numprint{1.474}$, with one instance improving from $\numprint{22.6}$\,s to $\numprint{0.58}$\,s. The dense families slow down by about five percent because they begin with a weaker upper bound.}

{\looseness=-1 (3) At $32$ threads, contraction rather than traversal dominates the remaining running time: measured per round, it accounts for about $80$ percent of the runtime and is largely sequential. The implementation partitions the work by vertices before contraction, so different threads can map different vertices to the same contracted vertex and generate the same contracted edge. Combining the weights of these duplicate edges requires a concurrent hash table keyed by the endpoint pair, while constructing the contracted graph from the table inserts edges one at a time inside a critical section, serializing this step. Partitioning instead by contracted vertex removes both bottlenecks: each thread owns the neighbor lists of its assigned contracted vertices, so no two threads write to the same entry, and parallel edges are summed in a thread-private dense array that fits into cache once only a few thousand vertices remain. Since each contracted vertex forms one unit of work regardless of how many pre-contraction vertices it contains, this approach is used only when the groups are of comparable size. On one social-network instance, for example, the first round reduces $13$ million vertices to a few thousand, and one very large group would otherwise occupy a single thread for most of the round. Together, these changes yield a speedup of $\numprint{1.294}$.}

{Two smaller optimizations provide further gains. (4) The edge record is reduced from $40$ to $24$ bytes. Two of its five fields are used only by the maximum flow algorithms that \texttt{VieCut} also contains, which are never called by the exact minimum cut path, so they are compiled out of the minimum cut binaries. Since each CAPFOREST traversal scans every edge once per round, the smaller record reduces memory traffic and yields a speedup of $\numprint{1.131}$. (5) A third priority queue stores each bucket as an intrusive linked list and targets instances between the two existing queue variants, where the cut is small enough for buckets but most buckets remain empty, yielding a speedup of $\numprint{1.027}$.}

\paragraph*{Correctness.} \looseness=-1 Matching the baseline on all $53$ instances is validation, not proof. Some accepted changes preserve the published invariants~\cite{henzinger2018shared} by construction: the algorithm is correct for any valid upper bound $\hat\lambda$, the lower-bound argument holds however the passes divide the traversal, and the queue choice only reorders the search. The remaining changes are implementation-level and checked empirically, by the assertions of Section~\ref{s:agentic} and the identical cuts. A human read all $23$ accepted diffs and checked each of them for correctness.

\paragraph*{Tuning and the instance set.} The thresholds in the accepted changes, such as the heap selection and the small-graph routing at $\numprint{10000}$ vertices, were fitted on the same $53$ instances the speedups are reported on, since the methodology deliberately tunes for the workload it is given. The risk of overfitting is bounded: acceptance requires improving the geometric mean over the whole set, the accepted changes improve both the real-world $k$-cores and the DIMACS core instances, and the largest improvements remove asymptotic bottlenecks rather than fit constants. Users may re-tune the thresholds on their own workload; the effect on instances held out from tuning remains to be evaluated.

\section{Discussion}
\label{conclusion}

We introduced \emph{agentic algorithm engineering}, in which autonomous LLM agents run the algorithm engineering cycle under correctness assertions, and applied it to our extensively hand-tuned shared-memory parallel exact minimum cut algorithm. The accepted changes fall into three categories: the largest remove implicit assumptions of the original tuning that the DIMACS core instances violate, such as a priority queue sized by the cut value; a broad share is standard performance engineering that needs no insight into minimum cuts; and a few alter decisions rather than execution, such as probing whether the passes share the visited array. The agent did not invent a new algorithm, and the assertions fix its output to the baseline's cuts; the sessions show that the complete cycle can be run at a cost low enough to engineer regimes outside the original design envelope.
Running the process was not effortless. An agent that optimizes running time alone produces code that is fast but wrong and chases changes that help one instance while slowing the set, so the correctness assertions, the geometric mean over the whole set as the only acceptance criterion and a human review of every accepted diff were necessary; $21$ of the $44$ experiments were discarded. The context window had to be managed explicitly, with all session state on disk, and the benchmark was the bottleneck: every experiment costs a full run over $53$ instances, so a session is paced by machine time rather than by the agent. Future work includes applying the methodology to other well-engineered codes and to metrics beyond running time, such as the solution quality of heuristics or the memory footprint of an implementation, including \hbox{trade-offs between them.}

\paragraph*{Acknowledgements.} This research was funded in part by NSF grant numbers CCF-2109988, OAC-2402560, and CCF-2453324 (Bader), and by the Deutsche Forschungsgemeinschaft (DFG, German Research Foundation), project number 519626652 (Schulz).

\vfill \pagebreak
\bibliography{quellen}

\appendix

\vfill\pagebreak
\section{Detailed Related Work}
\label{app:related}

Ford and Fulkerson~\cite{ford1956maximal} proved that the minimum $s$-$t$-cut equals the maximum $s$-$t$-flow, and Gomory and Hu~\cite{gomory1961multi} observed that the global minimum cut can be computed with $n-1$ minimum $s$-$t$-cut computations. Hence, improved maximum flow algorithms such as push-relabel~\cite{goldberg1988new} were used to obtain better global minimum cut algorithms~\cite{karger1996new}. Hao and Orlin~\cite{hao1992faster} adapt push-relabel to pass information to future flow computations, achieving a total running time of $O(mn\log{\frac{n^2}{m}})$. Gabow~\cite{gabow1991matroid} gives an algorithm running in $O(m+\lambda^2n \log(n/\lambda))$.
Flow-free approaches contract edges instead. Padberg and Rinaldi~\cite{padberg1990efficient} give four local tests that identify edges whose contraction preserves at least one minimum cut; applying them exhaustively costs $O(nm)$ time. Chekuri~\etal\cite{Chekuri:1997:ESM:314161.314315} instead apply the tests in linear-time sweeps over the graph, which may miss some contractible edges, and repeat the sweeps only while they shrink the graph by a constant factor. This costs $O(m \log n)$ time in total and finds almost as many contractions in practice. Nagamochi~\etal\cite{nagamochi1992computing,nagamochi1994implementing} repeatedly use maximum spanning forests to find contractible edges, running in $O(mn+n^2\log{n})$. Stoer and Wagner~\cite{stoer1997simple} give a simpler variant of the same asymptotic complexity but significantly worse practical performance~\cite{junger2000practical}. Among algorithms with better bounds, Kawarabayashi and Thorup~\cite{kawarabayashi2015deterministic} give a deterministic $O(m \log^{12}{n})$ algorithm, later improved by Henzinger~\etal\cite{henzinger2017local} to $O(m \log^2{n} \log \log^2 n)$. Matula~\cite{matula1993linear} gives a linear time $(2+\varepsilon)$-approximation. Karger and Stein~\cite{karger1996new} give a randomized algorithm based on random edge contractions. The algorithms of Hao and Orlin, Nagamochi et~al., Padberg and Rinaldi, Stoer and Wagner, and Karger and Stein have all been implemented and compared experimentally~\cite{Chekuri:1997:ESM:314161.314315,junger2000practical,levine1997experimental}, whereas, to our knowledge, those of Kawarabayashi and Thorup and of Henzinger~\etal have not.

Gawrychowski~\etal\cite{gawrychowski2020minimum} improve the randomized bound of Karger for weighted graphs to $\Oh{m \log^2 n}$, and for simple graphs Ghaffari~\etal\cite{ghaffari2020faster} give an algorithm with running time $\Oh{m \log n}$ that uses random 2-out contractions.
Li and Panigrahi~\cite{li2020deterministic} introduce the \emph{isolating cuts} technique, Li~\cite{li2021deterministic} then gives a deterministic algorithm with running time $m^{1+o(1)}$, and Henzinger~\etal\cite{henzinger2024deterministic} give a deterministic $\tilde{\mathcal{O}}(m)$ algorithm for weighted graphs.
Anderson and Blelloch~\cite{anderson2023parallel} give a parallel algorithm with $\Oh{m \log^2 n}$ work and polylogarithmic depth, which matches the best sequential work bound.
None of the algorithms of this paragraph have been implemented, so their performance in practice is unknown. Moreover, to the best of our knowledge, the only distributed implementation for the problem is still the one of Gianinazzi~\etal\cite{gianinazzi2018communication}, and there is no GPU implementation of the problem.

\clearpage
\section{Pseudocode}
\label{app:pseudocode}

We give the pseudocode of the two routines described in Section~\ref{cuts}. Algorithm~\ref{algo:parnoi} is the parallel variant of CAPFOREST that finds contractible edges and Algorithm~\ref{algo:parmc} is the overall minimum cut algorithm. Both are taken from the original paper~\cite{henzinger2018shared}, which also contains the correctness proofs.

\begin{algorithm}[H]
  \begin{algorithmic}[1]
    \INPUT $G = (V,E,c) \leftarrow$ undirected graph $\hat\lambda \leftarrow$ upper bound for minimum cut, $\mathcal{T} \leftarrow$ shared array of vertex visits
    \OUTPUT $\mathcal{U} \leftarrow$ union-find data structure to mark contractible edges
    \State Label all vertices $v \in V$ ``unvisited'', blacklist $\mathcal{B}$ empty
    \State $\forall v \in V: r(v) \leftarrow 0$
    \State $\forall e \in E: q(e) \leftarrow 0$, $\alpha \leftarrow 0$
    \State $\mathcal{Q} \leftarrow$ empty priority queue
    \State Insert random vertex into $\mathcal{Q}$
    \While{$\mathcal{Q}$ not empty}
    \State $x \leftarrow \mathcal{Q}$.pop\_max() \Comment{Choose unvisited vertex with highest priority}
    \State Mark $x$ ``visited''
    \If{$\mathcal{T}(x) = \text{True}$} \Comment{Every vertex is visited only once}
    \State $\mathcal{B}(x) \leftarrow $ True
    \Else
    \State $\mathcal{T}(x) \leftarrow \text{True}$
    \State $\alpha \leftarrow \alpha + c(x) - 2 r(x)$

    \State $\hat\lambda \leftarrow min(\hat\lambda, \alpha)$
    \For{$e = \{x,y\} \leftarrow$ unscanned edge, where $y \not\in \mathcal{B}$}
    \If{$r(y) < \hat\lambda \leq r(y) + c(e)$}
    \State $\mathcal{U}$.union(x,y) \Comment{Mark edge $e$ to contract}
    \EndIf
    \State $r(y) \leftarrow r(y) + c(e)$
    \State $q(e) \leftarrow r(y)$
    \State $\mathcal{Q}(y) \leftarrow min(r(y), \hat\lambda)$
    \EndFor
    \EndIf
    \EndWhile

  \end{algorithmic}
  \caption{\label{algo:parnoi} Parallel CAPFOREST}
\end{algorithm}

\begin{algorithm}[H]
  \begin{algorithmic}[1]
    \INPUT $G = (V,E,c)$
    \State $\hat\lambda \leftarrow $ \texttt{VieCut}($G$), $G_C \leftarrow G$
    \While{$G_C$ has more than $2$ vertices}
    \State $\hat\lambda \leftarrow$ Parallel CAPFOREST($G_C,\hat\lambda$)
    \If{no edges marked contractible}
        \State  $\hat\lambda \leftarrow$ CAPFOREST($G_C,\hat\lambda$)
        \EndIf
        \State $G_C, \hat\lambda \leftarrow$ Parallel Graph Contract($G_C$)
        \EndWhile

        \State \Return $\hat\lambda$
  \end{algorithmic}
  \caption{\label{algo:parmc} Parallel Minimum Cut}

\end{algorithm}

\clearpage
\section{Instance Set}
\label{app:instances}

Table~\ref{tab:instances} lists the instances we evaluate on: families we
generate, the \emph{DIMACS core instances}, and $k$-cores of
real-world web and social graphs. For each one the table gives its size and its
minimum cut $\lambda$.

\subsection*{Generated Families: The DIMACS Core Instances}

The generated instances come from seven generators that reimplement the problem
families of Chekuri~\etal\cite{Chekuri:1997:ESM:314161.314315} and of
Levine~\cite{levine1997experimental}, and every family label in
Table~\ref{tab:instances} maps to its generator as follows: unions of random
cycles (\texttt{reg2}, \texttt{reg3} and \texttt{regsweep}, whose three
instances arise from sweeping the graph size, the seed and the cut value,
respectively), random graphs with heavy components (\texttt{noi1},
sweeping the size, and \texttt{noisweep}, sweeping the remaining parameters),
regular (\texttt{random}) and irregular (\texttt{irreg}) random
graphs, the generator of Padberg and Rinaldi~\cite{padberg1990efficient}
(\texttt{pr1} and \texttt{pr2}, its two types), bicycle wheels
(\texttt{bikewheel}) and pairs of interleaved cycles
(\texttt{dblcyc}). Those generators were never published, so ours follow
the descriptions given there. Each takes the number of vertices, the
parameters in the table and, where the generator is randomized, the seed
$s$ given in the table, and writes the METIS format; they are deterministic, so
the set is reproducible from the table alone rather than from stored files.

At the sizes the study states, our algorithm finishes in milliseconds, so we
enlarge every family until it takes seconds to minutes and keep the three
slowest instances of each. Memory, not time, sets how far this goes: where the
minimum cut is small the algorithm retains one contracted graph per round and
needs $\Theta(n\cdot m)$, which stops \texttt{reg2} at $n=\numprint{12800}$ and
\texttt{irreg} at $n=\numprint{64000}$; for \texttt{pr2} the cut itself grows
like $n^2$ and the priority queue is sized by it, so $\numprint{8000}$ vertices
already need hundreds of gigabytes. Regular random graphs are at the other
extreme and scale to $n=\numprint{1000000}$. These limits are limits of the
\emph{unmodified} implementation; the single-threaded session of
Section~\ref{ss:sessions} later removes exactly this per-round history and the
cut-sized queue. The set was fixed before the sessions, so the instance sizes,
and with them the reported factors, are capped by what the baseline could run
at all.

Bicycle wheels and interleaved cycles are the adversarial cases: the first give
all $n$ trivial cuts the same value, the second hide one cut of $2000$ below
$\Theta(n^2)$ cuts of value $2006$. Both defeat the tests of Padberg and
Rinaldi, leaving a contraction-based algorithm no edge it may contract cheaply,
and are therefore where such algorithms degrade first.

A dagger marks a minimum cut that the construction fixes in advance. Every other
value was confirmed by two independent implementations: the Nagamochi-Ibaraki
code of \texttt{VieCut} and, depending on the size, either a Stoer-Wagner
implementation that shares no code with it or the parallel exact algorithm of
Section~\ref{cuts}. They agree on every instance, and with the predicted values.

\subsection*{Real-World Graphs}

The second group are the web and social
graphs~\cite{benchmarksfornetworksanalysis,BRSLLP,BoVWFI} already used to
evaluate the shared-memory exact algorithm~\cite{henzinger2018shared}, built the
same way: as they are disconnected and full of low-degree vertices, we take four
$k$-cores~\cite{seidman1983network,batagelj2003m} of each whose minimum cut
differs from the minimum degree, and use the largest connected component. They
are far larger than the DIMACS core instances, up to $\numprint{68141228}$
vertices and $\numprint{3134185720}$ edges, yet are solved much faster: the
algorithm contracts them aggressively, while the DIMACS core instances are built so
that it cannot.

\begingroup
\small
\setlength{\tabcolsep}{5pt}
\setlength{\LTcapwidth}{\textwidth}
\begin{longtable}{llrrr}
\caption{The evaluation set. For every instance we give the number of
  vertices and edges and the value $\lambda$ of the minimum cut. A dagger marks
  a minimum cut that the construction of the family fixes in advance; every
  other value was computed. The upper block is generated (the DIMACS
  core instances), the lower block
  consists of $k$-cores of real-world graphs. For the generated block, the
  parameters include the seed $s$ where the generator is randomized, so
  each instance is reproduced by its row.}
\label{tab:instances}\\
\toprule
Instance & Parameters & $n$ & $m$ & $\lambda$\\
\midrule
\endfirsthead
\caption[]{The evaluation set \emph{(continued)}.}\\
\toprule
Instance & Parameters & $n$ & $m$ & $\lambda$\\
\midrule
\endhead
\bottomrule
\endlastfoot
\multicolumn{5}{l}{\emph{generated families (DIMACS core instances)}}\\*
\texttt{reg2} & $c=50$, $s=1$ & \numprint{3200} & \numprint{157664} & \numprint{100}$^{\dagger}$\\
\texttt{reg2} & $c=50$, $s=1$ & \numprint{6400} & \numprint{317541} & \numprint{100}$^{\dagger}$\\
\texttt{reg2} & $c=50$, $s=1$ & \numprint{12800} & \numprint{637542} & \numprint{100}$^{\dagger}$\\
\texttt{reg3} & $c=2$, $s=2$ & \numprint{65536} & \numprint{131071} & \numprint{4}$^{\dagger}$\\
\texttt{reg3} & $c=2$, $s=1$ & \numprint{65536} & \numprint{131072} & \numprint{4}$^{\dagger}$\\
\texttt{reg3} & $c=2$, $s=3$ & \numprint{65536} & \numprint{131068} & \numprint{4}$^{\dagger}$\\
\texttt{regsweep} & $c=32$, $s=1$ & \numprint{2000} & \numprint{63025} & \numprint{64}$^{\dagger}$\\
\texttt{regsweep} & $c=16$, $s=1$ & \numprint{2000} & \numprint{31783} & \numprint{32}$^{\dagger}$\\
\texttt{regsweep} & $c=8$, $s=1$ & \numprint{2000} & \numprint{15946} & \numprint{16}$^{\dagger}$\\
\texttt{noi1} & $d=20$, $k=4$, $P=100$, $s=1$ & \numprint{8000} & \numprint{6399200} & \numprint{1608874}\\
\texttt{noi1} & $d=20$, $k=4$, $P=100$, $s=1$ & \numprint{12000} & \numprint{14398800} & \numprint{2542217}\\
\texttt{noi1} & $d=10$, $k=4$, $P=100$, $s=1$ & \numprint{16000} & \numprint{12799200} & \numprint{1646880}\\
\texttt{noisweep} & $d=100$, $k=4$, $P=100$, $s=1$ & \numprint{4000} & \numprint{7998000} & \numprint{4761257}\\
\texttt{noisweep} & $d=20$, $k=4$, $P=1000$, $s=1$ & \numprint{4000} & \numprint{1599600} & \numprint{7423512}\\
\texttt{noisweep} & $d=20$, $k=1$, $P=100$, $s=1$ & \numprint{8000} & \numprint{6399200} & \numprint{7261322}\\
\texttt{random} & $d=32$, $s=1$ & \numprint{400000} & \numprint{6399736} & \numprint{30}\\
\texttt{random} & $d=16$, $s=1$ & \numprint{1000000} & \numprint{7999939} & \numprint{14}\\
\texttt{random} & $d=8$, $s=1$ & \numprint{1000000} & \numprint{3999989} & \numprint{6}\\
\texttt{irreg} & $\lambda=4$, $e=32000$, $s=1$ & \numprint{32000} & \numprint{95993} & \numprint{6}\\
\texttt{irreg} & $\lambda=4$, $e=64000$, $s=1$ & \numprint{64000} & \numprint{191995} & \numprint{6}\\
\texttt{irreg} & $\lambda=4$, $e=0$, $s=1$ & \numprint{64000} & \numprint{127999} & \numprint{4}$^{\dagger}$\\
\texttt{pr1} & type~1, $d=20$, $s=1$ & \numprint{8000} & \numprint{6400003} & \numprint{72454}\\
\texttt{pr1} & type~1, $d=20$, $s=1$ & \numprint{12000} & \numprint{14400216} & \numprint{110938}\\
\texttt{pr1} & type~1, $d=10$, $s=1$ & \numprint{16000} & \numprint{12807052} & \numprint{70611}\\
\texttt{pr2} & type~2, $d=50$, $s=1$ & \numprint{3000} & \numprint{2249760} & \numprint{56833860}\\
\texttt{pr2} & type~2, $d=20$, $s=1$ & \numprint{6000} & \numprint{3602216} & \numprint{91070487}\\
\texttt{pr2} & type~2, $d=20$, $s=1$ & \numprint{8000} & \numprint{6400003} & \numprint{161635880}\\
\texttt{bikewheel} & --- & \numprint{4096} & \numprint{8189} & \numprint{8188}$^{\dagger}$\\
\texttt{bikewheel} & --- & \numprint{16384} & \numprint{32765} & \numprint{32764}$^{\dagger}$\\
\texttt{bikewheel} & --- & \numprint{65536} & \numprint{131069} & \numprint{131068}$^{\dagger}$\\
\texttt{dblcyc} & --- & \numprint{4096} & \numprint{8192} & \numprint{2000}$^{\dagger}$\\
\texttt{dblcyc} & --- & \numprint{16384} & \numprint{32768} & \numprint{2000}$^{\dagger}$\\
\texttt{dblcyc} & --- & \numprint{65536} & \numprint{131072} & \numprint{2000}$^{\dagger}$\\
\addlinespace
\multicolumn{5}{l}{\emph{$k$-cores of real-world graphs}}\\*
\texttt{uk-2007-05} & $k=10$ & \numprint{68141228} & \numprint{3134185720} & \numprint{1}\\
\texttt{uk-2007-05} & $k=50$ & \numprint{16759375} & \numprint{1770324244} & \numprint{1}\\
\texttt{uk-2007-05} & $k=100$ & \numprint{3979369} & \numprint{869154056} & \numprint{1}\\
\texttt{uk-2007-05} & $k=1000$ & \numprint{223416} & \numprint{183373955} & \numprint{1}\\
\texttt{twitter-2010} & $k=25$ & \numprint{13083726} & \numprint{970195992} & \numprint{1}\\
\texttt{twitter-2010} & $k=30$ & \numprint{10469192} & \numprint{899618650} & \numprint{1}\\
\texttt{twitter-2010} & $k=50$ & \numprint{4417754} & \numprint{677688218} & \numprint{3}\\
\texttt{twitter-2010} & $k=60$ & \numprint{3529047} & \numprint{629640787} & \numprint{3}\\
\texttt{uk-2002} & $k=10$ & \numprint{9019566} & \numprint{225617173} & \numprint{1}\\
\texttt{uk-2002} & $k=30$ & \numprint{2553366} & \numprint{114612142} & \numprint{1}\\
\texttt{uk-2002} & $k=50$ & \numprint{783316} & \numprint{51882018} & \numprint{1}\\
\texttt{uk-2002} & $k=100$ & \numprint{98275} & \numprint{10689202} & \numprint{1}\\
\texttt{com-orkut} & $k=16$ & \numprint{2427486} & \numprint{112151730} & \numprint{14}\\
\texttt{com-orkut} & $k=95$ & \numprint{114190} & \numprint{17970565} & \numprint{89}\\
\texttt{com-orkut} & $k=98$ & \numprint{107486} & \numprint{17335612} & \numprint{76}\\
\texttt{com-orkut} & $k=100$ & \numprint{103911} & \numprint{16992328} & \numprint{70}\\
\texttt{hollywood-2011} & $k=20$ & \numprint{1319770} & \numprint{109132341} & \numprint{1}\\
\texttt{hollywood-2011} & $k=60$ & \numprint{576111} & \numprint{86835657} & \numprint{6}\\
\texttt{hollywood-2011} & $k=100$ & \numprint{328631} & \numprint{70765201} & \numprint{77}\\
\texttt{hollywood-2011} & $k=200$ & \numprint{138536} & \numprint{47103731} & \numprint{27}\\
\end{longtable}
\endgroup

\end{document}